\documentclass[a4paper,11pt]{article}

\usepackage{graphicx}  %%% for including graphics
\usepackage{url}       %%% for including URLs
\usepackage{times}
\usepackage{natbib}
\usepackage[margin=25mm]{geometry}
\usepackage{tikz}
\usetikzlibrary{matrix,positioning}
\usepackage{verbatim}
\usepackage{array}
\usepackage{multirow}
\usepackage{booktabs}
\mathchardef\mhyphen="2D % Define a "math hyphen"

\title{Re-Ranking Words to Improve Interpretability of Automatically Generated Topics}
\date{}
\author{\textbf{Areej Alokaili}\textsuperscript{1,2}, \textbf{Nikolaos Aletras}\textsuperscript{1} and \textbf{Mark Stevenson}\textsuperscript{1}\\
      \textsuperscript{1}University of Sheffield, United Kingdom\\
      \textsuperscript{2}King Saud University, Saudi Arabia\\
      {\texttt{\{areej.okaili,n.aletras,mark.stevenson\}@sheffield.ac.uk}}
}

\begin{document}
\maketitle

\thispagestyle{empty}
\pagestyle{empty}
\begin{abstract}
 Topics models, such as LDA, are widely used in Natural Language Processing. Making their output interpretable is an important area of research with applications to areas such as the enhancement of exploratory search interfaces and the development of interpretable machine learning models. Conventionally, topics are represented by their {\it n} most probable words, however, these representations are often difficult for humans to interpret. This paper explores the re-ranking of topic words to generate more interpretable topic representations. A range of approaches are compared and evaluated in two experiments. The first uses crowdworkers to associate topics represented by different word rankings with related documents. The second experiment is an automatic approach based on a document retrieval task applied on multiple domains. Results in both experiments demonstrate that re-ranking words improves topic interpretability and that the most effective re-ranking schemes were those which combine information about the importance of words both within topics and their relative frequency in the entire corpus.  In addition, close correlation between the results of the two evaluation approaches suggests that the automatic method proposed here could be used to evaluate re-ranking methods without the need for human judgements.
\end{abstract}
%%%%%%%%%%%%%%%%%%%%%%%%%%%%%%%%%%%%%%
\section{Introduction}
Probabilistic topic modelling \citep{Blei2012ProbabilisticModels} is a widely used approach in Natural Language Processing \citep{Boyd-Graber2017ApplicationsModels} with applications to areas such as enhancing exploratory search interfaces \citep{Chaney2012VisualizingModels,Aletras2014jcdl,Smith2017EvaluatingLabels,Aletras2017jasist} and developing interpretable machine learning models \citep{Paul2016InterpretableModeling}. A topic model, e.g. Latent Dirichlet Allocation (LDA) \citep{Blei2003LatentAllocation} learns a low-dimensional representation of documents as a mixture of latent variables called topics. Topics are multinomial distributions over a predefined vocabulary of words. 

Traditionally, topics have been represented by lists of the topic's $n$ most probable words, however it is not always straightforward to interpret them due to noisy or domain specific data, spurious word co-occurrences and highly-frequent/low-informative words assigned with high probability \citep{Chang2009ReadingModels.}.

\renewcommand{\arraystretch}{1}
\begin{table*}[!h] 

\caption{Examples of topics represented by 30 most probable words from New York Times. Less informative words are shown in {\bf bold}.}
\label{t:topic_examples30words}
\normalsize
\resizebox{\textwidth}{!}{
\begin{tabular}{>{\raggedright\arraybackslash}p{18cm}}%>{\centering\arraybackslash}p{2cm}} 
\toprule
\toprule
{\bf Topic}\\
\hline %& {\bf Prob. Mass} \\ \hline
space museum \textbf{years} history science earth mission \textbf{could} art shuttle universe flight  \textbf{people} \textbf{theory} \textbf{world}  radar crew  \textbf{site}  pincus   plane  \textbf{three}  scientists  \textbf{day} \textbf{century}  pilot   exhibit  \textbf{back} \textbf{anniversary}  landing \textbf{project}\\  
\hline

\textbf{percent} \textbf{million}  market company stock  \textbf{billion} \textbf{sales}  bank  shares  \textbf{price} \textbf{business}  investors  \textbf{money} share  \textbf{companies} \textbf{rates}  fund  \textbf{interest} \textbf{rate} \textbf{quarter} \textbf{prices}  investment  funds  financial amp analysts \textbf{growth} \textbf{industry} \textbf{york}  banks \\ 
\hline

 film  \textbf{even} movie  \textbf{world} stars  \textbf{man} \textbf{much} \textbf{little} \textbf{story} \textbf{good} \textbf{way}  star  \textbf{best} show  \textbf{see} \textbf{well} \textbf{seems} \textbf{american} \textbf{people} \textbf{love} hollywood   director  \textbf{big} \textbf{ever} \textbf{rating} \textbf{though} \textbf{great} \textbf{seem} production  \textbf{makes} \\ 
\hline

\textbf{officials} \textbf{agency} \textbf{office} \textbf{report} \textbf{department}  investigation    government  \textbf{former}  federal  \textbf{charges} \textbf{secret} \textbf{information} \textbf{card}  cia   law agents security  \textbf{documents} \textbf{case}  investigators  \textbf{official}  fraud   intelligence   illegal \textbf{commission} \textbf{service} police  \textbf{cards}  enforcement   attorney  \\

\bottomrule
\bottomrule
\end{tabular}}

\end{table*}

 %\citet{Bhatia2018}  showed that the topic model quality measures using topic-level methods can differ from the quality measures through a document-level approach and proposed a method for automatically predicting topic model quality. They introduced the first attempt to automate the identification of intruder topics which was introduced by ref {Chang et al. 2009} to estimate the quality of the model's document-topic allocation. They suggest the used of their automatic measure alongside topic coherence to provide a comprehensive topic model evaluation.

\par Improving the interpretability of topic models is an important area of research. A range of approaches have been developed including computing topic coherence \citep{Newman2010AutomaticCoherence,Mimno2011OptimizingModels, Aletras2013EvaluatingSemantics, Lau2014MachineQuality.}, determining optimal topic cardinality \citep{Lau2016TheCardinality}, labelling topics text and/or images \citep{lau2011automaticModels,Aletras2014acl,Aletras2013RepresentingImages,Aletras2017ecir,Sorodoc2017} and corpus pre-processing \citep{Schofield2017PullingModels}. However, methods for re-ranking topic words to improve topic interpretability have not been systematically evaluated yet. We hypothesise that some words relevant to a particular topic have not been assigned with a high probability due to data sparseness or low frequency in the training corpus. Our goal is to identify these words and re-rank the list representing the topic to make it more comprehensible. Table \ref{t:topic_examples30words} shows topics represented by the 30 most probable words. Words displayed in bold font are more general (less informative, e.g. with high document frequency) while the remaining words are more likely to represent a coherent thematic subject. For example in the second topic, relevant words (e.g. investment, fund) have been assigned with lower probability compared to less informative words (e.g. percent, million). As a result, these words will not appear in the top 10 words.

\par This paper compares several word ranking methods and evaluate them using two approaches. The first approach is based on a crowdsourcing task in which participants are provided with a document and a list of topics then asked to identify the correct one, i.e. the topic that is most closely associated with the document. Topics are represented by word lists ranked using different methods. The effectiveness of the re-ranking approaches is evaluated by computing the accuracy of the participants on identifying the correct topic. The second evaluation approach is based on an information retrieval (IR) task and does not rely on human judgements. The re-ranked words are used to form a query and retrieve a set of documents from the collection. The effectiveness of the word re-ranking is then evaluated in terms of how well it can retrieve documents in the collection related to the topic. Results show that re-ranking topic words improves performance in both experiments. 

% Contributions
\par The paper makes the following contributions. It highlights the problem of re-ranking topic words and demonstrates that it can improve topic interpretability. It introduces the first systematic evaluation of topic word re-ranking methods using two approaches: one based on crowdsourcing and another based on an IR task. The latter evaluation is an automated approach and does not rely on human judgments. Experiments demonstrate strong agreement between the results produced by these approaches which indicates that the IR-based approach could be used as an automated evaluation method in future studies. The paper also compares multiple approaches to word re-ranking and concludes that the most effective ones are those which combine information about the importance of words within topics and their relative frequency across the entire corpus. Code used in the experiments described in this paper can be downloaded from \url{https://github.com/areejokaili/topic_reranking}.

%%%%%%%%%%%%%%%%%%%%%%%%%%%%%%%%%%%
\section{Background}
\label{related_work}
%\par In this section, we briefly review the related work in re-ranking topic words generated by topic models. 

\par The standard approach to representing topics has been to show the top $n$ words with the highest probability given the topic, e.g.  \citep{Blei2009TOPICMODELS,Blei2009VisualizingExpressions}. However, these words may not be the ones that are most informative about the topic and a range of approaches to re-ranking them has been proposed in the literature. 

% tf-idf type measure 
\par \citet{Blei2009TOPICMODELS} proposed a re-ranking method inspired by the tf-idf word weighting which includes two types of information: firstly, the probability of a word given a topic of interest and, secondly, the same probability normalised by the average probability across all topics. The intuition behind this approach is that good words for representing a topic will be those which have both high probability for a given topic and low probability across all topics. \citet{Blei2009TOPICMODELS} did not describe any empirical evaluation of the effectiveness of their approach. 

\par Other word re-ranking methods have also combined information about the overall probability of a word and its relative probability in one topic compared to others. \citet{Chuang2012Termite:Models} describe a word re-ranking method applied within a topic model visualisation system. Their approach combines information about the word's overall probability within the corpus and its distinctiveness for a particular topic which is computed as the Kullback-Leibler divergence between the distribution of topics given the word and the distribution of topics. \citet{Sievert2014LDAvis:Topics} also combine both types of information within a topic visualisation system. 
% they combined information using a weighted sum while other approaches
% Applied to labelled corpora organised hierarchically? 
\citet{Bischof2012SummarizingExclusivity} developed an approach for hierarchical topic models which balances information about the word frequency in a topic and the exclusivity of that word to that topic relative to a set of similar topics within the hierarchy.

\par Others have proposed approaches that only take into account the relative probability of each word in a topic compared to the others. \citet{song2009TopicModeling} introduced a word ranking method based on normalising the probability of a word in a topic with the sum of the probabilities of that word across all topics. They evaluated their method against two other methods, the topic model's default ranking and the approach proposed by \citet{Blei2009TOPICMODELS}, and found that it performed better than either. A similar method was proposed by \citet{Taddy2012OnModels} who used the ratio of the probability of a word given a topic and the word's probability across the entire document collection.

%% Use information *during* inference
\par Recently, \citet{Xing2018DiagnosingVariability} proposed to use information gathered while fitting the topic model. They made use of topic parameters from posterior samples generated during Gibbs sampling and re-weighted words based on their variability. Words with high uncertainty (i.e. their probabilities fluctuate relatively highly) are less likely to be representative of the topic than those with more stable probability estimates.

% They incorporated information from the topic posterior distribution from multiple topic model samples to re-weight words. %At at topic level, they introduced a measure they call Topic Stability which measures the variability of a topic in the posterior in multiple models. 
% Words re-weighted based on their uncertainty within individual topics. Words with high uncertainty (i.e. their probability fluctuate highly) are less likely to be salient and representative of the topic and therefore, down weighted. 

% TODO: It's not clear to me where the term re-ranking part is in this work. Do we 
% need to include it? If we do the approach to term re-ranking they used should be 
% made clearer. [Clarified the ranking part of their approach ]
\par Topic re-ranking has also been explored within the context of measuring topic quality 
\citep{Gollapalli2018UsingLDA}. A main claim of that work is that word importance should not only depend on its probability within a topic but also on its association with relevant neighbour words in the corpus. This information is incorporated by constructing topic-specific graphs capturing neighborhood words in a corpus. The PageRank \citep{Brin1998TheEngine} algorithm is used to assign word importance scores based on centrality and then re-rank words based on their importance. The top $n$ words with the highest PageRank values are used to compute the topics quality. %which are then used to measure the topics quality. Topic words are to re-rank words based on their PageRank value %They evaluated their measures on classification and intruder word identification tasks and found that their measures always achieve Superior performance. 

\par A common characteristic of previous work on topic word re-ranking is that it has been carried out within the context of an application of topic models (e.g. topic visualisation) and approaches have been evaluated in terms of these applications, if at all. The fact that word re-ranking methods have been considered in previous studies demonstrates their importance. The lack of direct and systematic evaluation is addressed in this work.

\section{Word Re-ranking Methods}
\label{rank}
\par This paper explores a range of methods for word re-ranking based around the main approaches that have been applied to the problem (see Section \ref{related_work}). Let $\hat{\varphi}_{w, t}$ be the probability of a word $w$ given a topic $t$ produced by a topic model, e.g. LDA.\footnote{We use the topic-word posterior distribution of LDA, but re-ranking can be applied to any topic model that estimates probabilities for words given topics.} The following methods are used to re-rank topic words.

\paragraph{Original LDA Ranking ($\mathbf{R_{Orig}}$)} 
The most obvious and commonly used method for ranking words associated with a topic is to use $\hat{\varphi}_{w, t}$ to score each word, i.e. $score_{w, t} = \hat{\varphi}_{w, t}$. The ranking generated by this scoring function is equivalent to choosing the $n$ most probable words for the topic and is referred to as $R_{Orig}$. 

\paragraph{Normalised LDA Ranking ($\mathbf{R_{Norm}}$)} 
The first re-ranking method is a simple extension of $R_{Orig}$ that represents approaches that normalise the probability of a word given a particular topic by the sum of probabilities for that word across all topics \citep{song2009TopicModeling,Taddy2012OnModels}. This measure is computed as: 
% Song et al. \citep{song2009TopicModeling} score words by normalising the marginal probability of a term given a particular topic by the sum of probabilities for that words across all topics in the model:

\begin{equation}
\label{eq-song2009}
score_{w, t}=\frac {\hat{\varphi}_{w, t}}{\displaystyle\sum_{j=1}^T \hat{\varphi}_{w, j}}
\end{equation}
where $T$ denotes the number of topics in the model. This approach scales the importance of words based on their overall occurrence within all topics in the model and downweights those that occur frequently. 

%The approach proposed by Song et al. \citep{song2009TopicModeling} was chosen because it appears to compare well against $R_{Orig}$ and $R_{TFIDF}$.  
% Song et al. \citep{song2009TopicModeling} compared $R_{Orig}$, $R_{TFIDF}$ and $R_{Norm}$ using two tasks in which annotators were asked to judge whether words included in topic representations were relevant or not. They found that $R_{Norm}$ performed best. 

\paragraph{Tf-idf Ranking ($\mathbf{R_{TFIDF}}$)}%{\bf $R_{TFIDF}$}: 
The second re-ranking method was proposed by \citet{Blei2009TOPICMODELS} and represents methods that combine information about the probability of a word in a single topic with information about its probability across all topics \citep{Bischof2012SummarizingExclusivity,Chuang2012Termite:Models,Sievert2014LDAvis:Topics}. Blei and Lafferty re-rank each word as:

% proposed an alternative ranking method inspired by the \texttt{tf-idf} term weighting from Information Retrieval \citep{Baeza-Yates1999ModernRetrieval}:

\begin{equation}
\label{eq-2}
{score_{w, t}}= \hat{\varphi}_{w, t} \; log  \frac{\hat{\varphi}_{w, t}} {\left(\displaystyle\prod_{j=1}^T\hat{\varphi}_{w, j}\right)^ \frac{1}{T}}
\end{equation}
%
% The second term scales the importance of words based on their overall occurrence within all topics in the model and downweights those that occur frequently.  
% Blei and Lafferty \citep{Blei2009TOPICMODELS} prefer this method to the default ranking but did not describe any empirical evaluation of its effectiveness. 

\paragraph{Inverse Document Frequency (IDF) Ranking ($\mathbf{R_{IDF}}$)}%$R_{IDF}$: 
The final word re-ranking method explored in this paper is a variant on the previous method that takes account of a word's distribution across documents rather than topics. This method has not been explored in previous literature. In this approach each word is weighted by the Inverse Document Frequency (IDF) score across the corpus used to train the topic model: 
\begin{equation}
\label{eq-3}
{score_{w,t}} = \hat{\varphi}_{w, t} \; {log\frac{|D|}{|D_w|} }
\end{equation} 
where $D$ is the entire document collection and $D_{w}$ the documents within $D$ containing the word $w$.

\par To better understand the effect of re-ranking words, consider the various representations of two topics shown in Table \ref{topics_example}. The first row for each topic represents the baseline rank produced by topic model ($R_{Orig}$), while the other rows show the topic after re-ranking using Equations  \ref{eq-song2009}, \ref{eq-2} and \ref{eq-3}, respectively. The bold words included in the original ranking ($R_{Orig}$) are down weighted and removed by at least two methods. Underlined words are those weighted higher by a re-ranking method and included in the topic representation. 

\renewcommand{\arraystretch}{1}
 \begin{table*} [!h]
\caption{Examples of topic representations produced using various ranking approaches. Words in the ${R}_{Orig}$ representation that were removed by at least two methods are shown in {\bf bold}. Words that are ranked higher by the other approaches and included in the topic representation are shown \underline{underlined}.}
\label{topics_example}
\centering
\normalsize
\begin{tabular}{ll}
\toprule
Method & Topic \\

\midrule

${R}_{Orig}$ & space museum \textbf{years} history science earth mission \textbf{could} art shuttle\\
$R_{Norm}$ & \underline{pincus} \underline{abrams} \underline{downey} \underline{gettysburg} \underline{particles} \underline{sims} \underline{emery} \underline{landers} \underline{lillian} \underline{alamo} \\
${R}_{TFIDF}$  &  museum space earth shuttle \underline{pincus} science \underline{universe} \underline{radar} \underline{exhibit} art \\
${R}_{IDF}$ & museum space \underline{pincus} science earth shuttle \underline{universe} \underline{radar} history mission\\
\midrule
${R}_{Orig}$ & film \textbf{even} movie \textbf{world} stars \textbf{man} \textbf{much} \textbf{little} \textbf{story} \textbf{good}  \\
$R_{Norm}$ & \underline{vampire} \underline{que} \underline{winchell} \underline{tomei} \underline{westin} \underline{swain} \underline{marisa} \underline{laughlin} \underline{faye} \underline{beatty}\\
${R}_{TFIDF}$  & film movie stars \underline{vampire} \underline{rating} \underline{spielberg} \underline{hollywood} \underline{star} \underline{characters} \underline{actors}\\
${R}_{IDF}$ & film movie stars \underline{vampire} \underline{rating} \underline{star} \underline{spielberg} even \underline{hollywood} story \\
\bottomrule

\end{tabular}
% }
\end{table*}

%%%%%%%%%%%%%%%%%%%%%%%%%%%%%%%%%%%%

\section{Experiment 1: Human Evaluation of Topic Interpretability}
\label{experiment1}
\par The first experiment compares the effectiveness of different topic representations (i.e. word re-rankings) by asking humans to choose the correct topic for a given document. We hypothesize that humans would be able to find the correct topic more easily when the representation is more interpretable.  %the information they deliver for the document collection at a document-level. 

\subsection{Dataset and Preprocessing}\label{sec:data}
\par We randomly sampled approximately 33,000 news articles from the New York Times included in the English GigaWord corpus fifth edition.\footnote{\url{https://catalog.ldc.upenn.edu/LDC2011T07}} Documents were tokenized and stopwords removed. Words occurring in fewer than five or more than half of the documents were also removed to control for rare and common words. The size of the resulting vocabulary is approximately 52,000 words.
%Standard pre-processing steps were performed including tokenization, stopwords removal and filtering the data from rare and common words based on document frequency, which covers any words that appear in less than five documents or appear in more than half of the documents.

\subsection{Topic Generation}
\par Topics were generated using LDA's implementation in Gensim\footnote{\url{https://radimrehurek.com/gensim}} fitted with online variational Bayes \citep{Hoffman2010OnlineAllocation}. The most important tuning parameter for LDA models is the number of topics and it was set to 50 after experimenting with varying number of topics optimised for coherence. To assess the quality of the resulting LDA models, topic coherence was computed\footnote{The implementations available in Gensim were used.} using: (1) $C_{V}$ \citep{Roder2015ExploringMeasures}; (2) $C_{UCI}$ \citep{Newman2010AutomaticCoherence}; and (3) $C_{NPMI}$ \citep{Bouma2009}.

\subsection{Crowdsourcing Task}

A job was created on the Figure Eight\footnote{ \url{https://www.figure-eight.com/}} crowdsourcing platform (previously known as CrowdFlower) in which participants were presented with ten micro-tasks per page\footnote{One of the ten micro-tasks on each page is reserved for quality assessment.}. Each micro-task consists of a text followed by six topics represented by a list of $n$ words selected using one of the re-ranking methods presented in Section~\ref{rank}. Participants were asked to select the topic that was most closely associated with the text. 

\par Micro-tasks were created using 48 New York Times articles (see Section \ref{sec:data}). The correct answer is the topic with the highest probability given the article and incorrect answers (i.e. distractors)  are five topics with low probabilities. The probability of the correct topic was at least 0.6 and the probability of the five distractors lower than 0.3.\footnote{Alternative values for these parameters were explored but it was found that lowering the probability of the correct answer and/or raising the probability of the distractors made the task too difficult.}
% we found that the task become too difficult for participants when there was less distinction between the correct and alternative topics.} 
Each article micro-tasks were created using each of the four ranking methods (Section \ref{rank})  generated from topics created using three cardinalities (5, 10, 20). Five assessments were obtained for each micro-task and consequently 60 judgments were obtained for each document.\footnote{ 4 (ranking methods) $\times$ 3 (cardinalities) $\times$ 5 (judgments per document)}

\begin{figure*} [ht]
\centering
\includegraphics[width=0.9\textwidth]{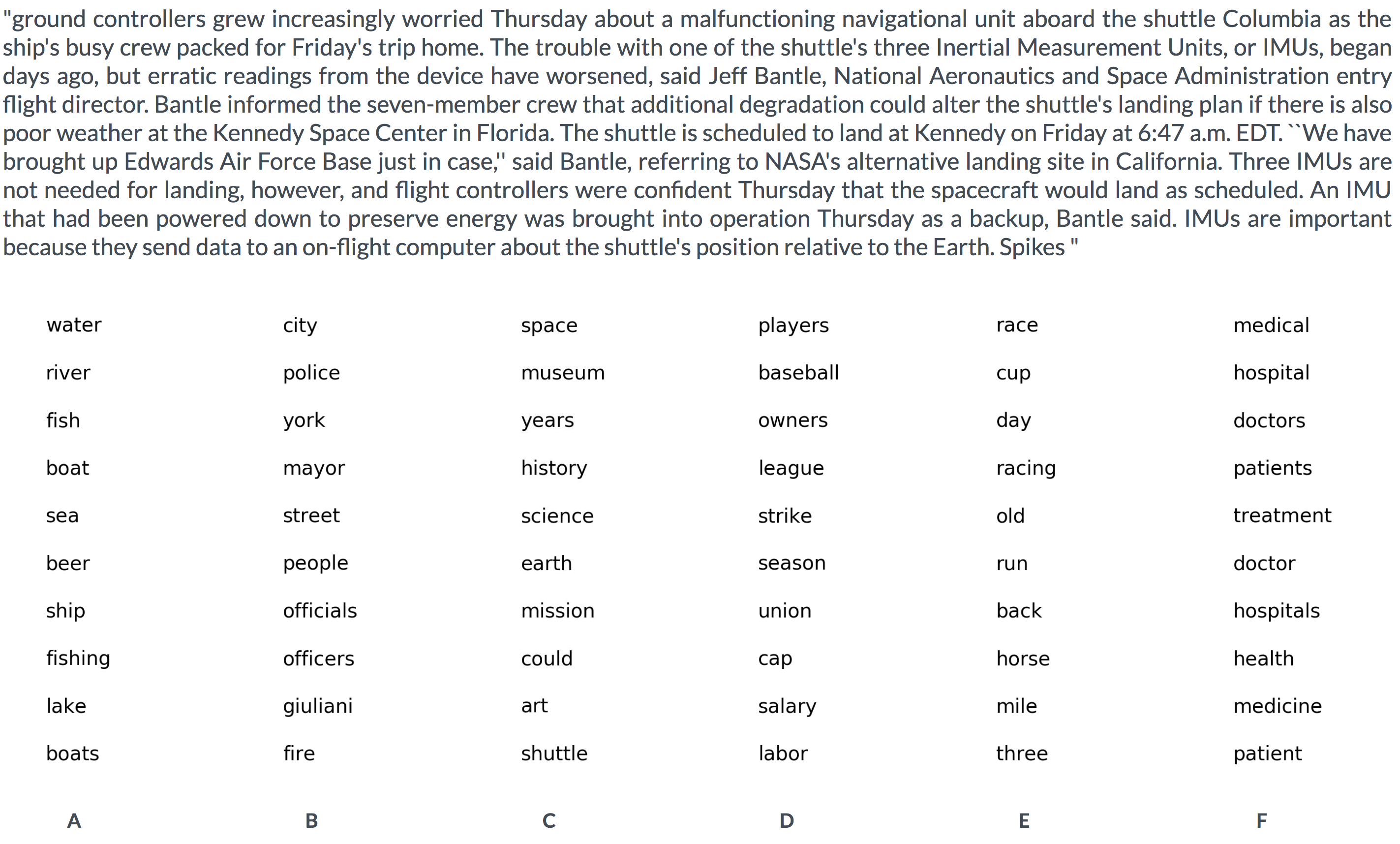}
\caption{Example of the crowdsourcing micro-task interface.} \label{fig:task}
\end{figure*} 

\par % Participants are asked to find the topic that is most closely associated with the document. 
Figure \ref{fig:task} shows an example of the micro-task presented to participants were they are asked to choose one of the topics.  Participants are first provided with a brief description and an example to help them understand the task, followed by a quiz of ten micro-tasks to ensure their reliability and to eliminate random answers \citep{Kazai2011InEvaluation}. Participants who fail to answer seven out of ten micro-tasks correctly are eliminated from the job. If they qualify and proceed, a further quality micro-task is added per page and they need to maintain an accuracy of responses above 70\%. 
% \textcolor{blue}{In total, 2,880 micro-tasks are in the study}. 
To ensure non-redundant results, participants were always shown questions using the same topic word re-ranking method and the same number of words per topic. Also, participants can only answer a single page of 10 micro-tasks. % explain in figure 1 where the topic shown use the same exact rankings

\subsection{Results and Discussion}
\label{results}

Results for $R_{Orig}$, $R_{Norm}$,  $R_{TFIDF}$ and $R_{IDF}$ when topics are represented by the 5, 10 and 20 highest scoring words are shown in Table \ref{result1}. \emph{Accuracy} represents the percentage of questions for which participants were able to identify the correct topic (i.e. topic with the highest probability given the document). \emph{Time/page} is the mean time taken for participants to complete a page of 10 questions. 
\emph{Coherence} is the average coherence of the topics, computed using NPMI \citep{Aletras2013EvaluatingSemantics}\footnote{The implementation provided by  \url{https://github.com/jhlau/topic_interpretability} was used.}.

\begin{table}[!htb]
\centering 
\caption{Results of experiment comparing re-ranking methods in which crowdsourcing participants were asked to associate topic representations with documents. Topics are represented with their top 5, 10 or 20 probable words.}
\label{result1}
\resizebox{0.6\textwidth}{!}{%
\begin{tabular}{c|ccccc}
\toprule 
\multirow{2}{*}{\#words} & &   \multicolumn{4}{c}{Ranking Methods} \\ 
 &   & {\bfseries}  ${R}_{Orig}$ & $R_{Norm}$ & ${R}_{TFIDF}$   & ${R}_{IDF}$\\  
\midrule
\multirow{2}{*}{5}& Accuracy (\%) & 64  & 55 &  70  & {\bf 73} \\  

& Time/page & 11:46 & 13:13 & 13:00   & 11:28 \\ 

& Coherence (NPMI) & 0.092 & 0.035 &	0.112 	 & 0.100 \\ 
\midrule
\multirow{2}{*}{10}& Accuracy (\%) &  67   & 48 & {\bf 76}  &  70 \\ 

& Time/page &   12:40 &  15:23 &  12:15  & 12:31\\

& Coherence (NPMI) & 0.072	& 0.038 & 0.091	 & 0.084 \\
\midrule
\multirow{2}{*}{20} & Accuracy (\%) & 69 & 64  &  {\bf 74} & 72 \\

& Time/page &   14:18& 14:30& 11:47  & 11:13 \\

 & Coherence (NPMI) & 0.050	&	0.029 & 0.071  &0.062  \\ 
\midrule 
\end{tabular}
}

\end{table}

Results show a variation in performance which indicates that re-ranking words affects individual's ability to interpret topics. Performance when the words are ranked using $R_{TFIDF}$ and $R_{IDF}$ outperform the default ranking ($R_{Orig}$). Performance when words are ranked using $R_{Norm}$ is considerably lower than their word re-ranking methods, both in terms of accuracy and time taken to complete the task. 

\par These results show that the improvement obtained by using $R_{TFIDF}$ and $R_{IDF}$ is consistent when the number of words in the representation is varied. Results using $R_{Orig}$ improve as the number of words increases but never achieve the same performance as the re-ranking methods (except $R_{Norm}$), even when 20 words are included. This demonstrates that choosing the most appropriate words to represent a topic is more useful than simply increasing the number of words shown to the user. In fact, increasing the number of words shown for the default ranking appears to come at the cost of slowing down the time taken for a user to interpret the topic. The same increase in task completion time is not observed for $R_{TFIDF}$ and $R_{IDF}$ and this may be down to the fact that more useful words appear earlier in the ranking, allowing participants to interpret the topic more quickly. 

The $R_{TFIDF}$ and $R_{IDF}$ approaches both combine information about the word's importance within an individual topic and across the entire document collection which results into more effective rankings than $R_{Orig}$. On the other hand, $R_{Norm}$ only considers the relative importance of a word across topics and it would be possible for a word with a relatively low probability given the topic to be ranked highly if that word also had low probability across all the other topics. 

Our results contrast with those reported by \citet{song2009TopicModeling} who concluded that $R_{Norm}$ was more effective for word re-ranking than $R_{Orig}$ and $R_{TFIDF}$ (see Section \ref{rank}). However, their evaluation methodology used a single annotator per-task and asked them to judge whether words included within topic representations were important or not. Our approach measures a participants ability to interpret topic representations more directly and makes use of multiple annotations. The low results for $R_{Norm}$ suggest that crowdworkers were simply unable to interpret many of the topics and, in those cases, their judgments about which words are important are unlikely to be reliable. 

\par Overall $R_{TFIDF}$ appears to be the most effective of the re-ranking approaches evaluated. This method achieves the best performance for 10 and 20 words, although not as well as $R_{IDF}$ for 5 words. 

%%%%%%%%%%%%%%%%%%%%%
\section{Experiment 2: Automatic Evaluation of Topic Interpretability via Document Retrieval}
\label{experiment2}
\par In this second experiment, we automate the evaluation of the different topic representations obtained by re-ranking the topic words. The automated evaluation is based on an IR task in which the re-ranked topic words are used to form a query and retrieve documents relevant to the topic. The motivation behind this approach is that the most effective re-rankings are the ones that can retrieve documents related to the topic, while ineffective re-rankings will not be able to distinguish these from other documents in the collection. This evaluation method does not rely on human judgments, unlike the crowdsourcing approach presented in the previous section.
%has the advantage that it can be easily automated and is also based on a final task.  

\subsection{Evaluation Pipeline}
\label{eval_pipeline}
% Mark: I've re-ordered this section so that a general description of the evaluation approach is presented first. Ideally it would be explained more formally but that will have to wait.  

The evaluation approach assumes that given a document collection in which each document is mapped to a label (or labels) indicating its topic. We refer to these labels as {\it gold standard topics} (to distinguish them from the automatically generated topics created by the topic model). 

First, a set of automatically generated topics are created by running a topic model over a document collection. For each gold standard topic, a set of all documents labelled with that topic is created. The document-topic distribution created by the topic model  is then used to identify the most probable automatically generated topic within that set of documents. This is achieved by summing the document-topic distributions and choosing the automatically generated topic with the highest value. A query is then created by selecting the re-ranked top $n$ words from that automatically generated topic and use it to retrieve a set of documents from the collection. The set of retrieved documents is then compared against the set of all documents labelled with the gold standard label.

% \subsection{Dataset and Preprocessing}
\subsection{Datasets}\label{sec:ir_data}

Evaluation was carried out using datasets representing documents from a wide range of domains: news articles, scientific literature and online reviews. 

\subsubsection{New York Times}
\par  A subset of the NYT annotated dataset\footnote{\url{ https://catalog.ldc.upenn.edu/LDC2008T19}} consisting of approximately approximately 39,000 articles was used for this experiment\footnote{Note that this is a different dataset to the one used for experiment 1 and contains the gold standard topics required for this evaluation}.  This collection contains news articles from the {\it New York Times} labelled with 1,746 topics which we use as gold standard labels. These labels, which we refer to as NYT\_topics, belong to a controlled set of topic categories and have been manually verified by NYTimes.com production staff. Each article has at least one NYT\_topic, and articles are organised into a topic hierarchy. Examples of NYT\_topics include: 
\begin{itemize}
%% \item \textit{Top/Features/Travel/Guides/Destinations/North America/United
%States/Arizona}
\item \textit{Top/Features/Travel/Guides/Destinations/North America/United States}
\item \textit{Top/News/New York and Region}
\item \textit{Top/News/Technology}
\end{itemize}

The hierarchy into which the topics are organised is quite deep in some places and consequently we truncated each topic to the top most four levels of the hierarchy to control the number of topics. For example, the topic \textit{Top/Features/Travel/Guides/ Destinations/North America/United States} is truncated to \textit{Top/Features/Travel/Guides}. This produces a total of 132 truncated NYT\_topics. The number of articles associated with each of the 132 NYT\_topics ranges from 1 to 18,489. To avoid NYT\_topics that are associated with small numbers of documents, we used the 50 NYT\_topics that are associated with the most documents which resulted in NYT\_topic that are each associated with at least 560 documents.

\subsubsection{MEDLINE} 
\par 

MEDLINE contains abstracts of more than 25 million scientific publications in medicine and related fields. These abstracts are labelled with \textit{Medical Subject Headings} (MeSH) codes which index publications into a hierarchy structure. Each publication is associated with a set of MeSH codes to describe the content of the publication. 
% \footnote{\url{https://www.nlm.nih.gov/databases/download/pubmed\_medline.html}} is a primary medical literature resource with more than 25 million records covering publications in biomedicine and health from 1805 to present. MEDLINE uses more than 19,000 \textit{Medical Subject Headings} \textbf{(MeSH)} to index and catalog publications into a hierarchy structure. Each publication is associated with a set of Mesh codes to describe the content of the publication. 
\par The 50 most frequently used MeSH codes with the most publications from a subset of MEDLINE containing publications from 2017. This set of code are referred to as \textit{MeSH\_topics}. 
% of the 2017 MEDLINE annual baseline database (partition files from pubmed18n0892 - pubmed18n0928). The extracted MeSH codes will be refered to as \textit{MeSH\_topics} which will be used to identify all the publications that belong to them. 
%%%%%%%%%%%%%%%%%%%
\subsubsection{Amazon Product Reviews}

\par The Amazon Product Reviews dataset \citep{McAuley2015}\footnote{\url{http://jmcauley.ucsd.edu/data/amazon}} contains reviews of products purchased from the Amazon website. Reviews are organized into 24 top-level categories each of which is divided into subcategories. The number of subcategories ranges from 1 to 1961.  
% and the number of  reviews in each category ranges from 1 to 22K. 
We chose eight main categories (\textit{Cell Phones and Accessories, Electronics, Movies and TV, Musical Instrument, Office Products, Pet Supplies, Tools and Home Improvement and Automotive}) and extracted the 10 sub-categories with the most reviews from each which yielded 76 distinct sub-categories. This set of categories is referred to as  \textit{AMZ\_topics}.
5,000 product reviews are extracted from each main category and reviews must belong to at least one category from the 50 most frequent in the AMZ\_topics, resulting in a total of 40,000 reviews.

\begin{table} 
\caption{Datasets statistics.}
\label{t:datasets_stat}

\centering
\resizebox{0.45\textwidth}{!}{%
\begin{tabular}{ccc} 
\hline
Dataset &  Documents & Distinct Words\\
\hline
NYT Annotated   & 39,218 & 60,339\\
MEDLINE   & 23,640 & 18,571 \\
Amazon   & 40,000 &  24,943\\

\hline
\end{tabular}
}
\end{table}

\color{black}
%%%%%%%%%%%%%%
\subsection{Experimental Settings}

\par Each of the datasets was indexed using Apache Lucene.\footnote{\url{ http://lucene.apache.org/}} The same preprocessing steps used in Experiment 1 were applied to the datasets and the statistics of the datasets are shown in Table \ref{t:datasets_stat}. 
\par For each dataset, LDA was used to generate topics and the number of topics for each dataset was set based on optimising for coherence which yielded 35 for NYT, 45 for MEDLINE and 35 for Amazon. The automatically generated topic that was most closely associated with each of the gold topics (i.e. 50 NYT\_topics, 50 MeSH\_topics and 50 AMZ\_topics) were identified by applying the process outlined above in Section \ref{eval_pipeline}. The top 5, 10 and 20 words from this topic is used to form a query which is submitted to Lucene. The BM25 retrieval model \citep{Robertson2004UnderstandingIDF} was used to measure the similarity between the document to a given query. The documents retrieved by applying these queries are compared against the entire set of documents labelled with the dataset topics (i.e. NYT\_topic, MeSH\_topics, or AMZ\_topics) by computing Mean Average Precision (MAP)\footnote{MAP is computed using the \textit{trec\_eval} tool:  \url{ http://trec.nist.gov/trec_eval/}} which is commonly used as a single metric to summarise IR system performance.

\subsection{Results and Discussion}
\label{resultsIR}

Queries were created using the top 5, 10, and 20 topic words using the $R_{Orig}$, $R_{Norm}$, $R_{TFIDF}$, and $R_{IDF}$ re-rankings and applied to each of the three datasets (Section \ref{sec:ir_data}). Results are shown in Table \ref{result3}.

\begin{table}[t!]
\centering
\caption{Results of experiment in which top 5, 10 and 20 ranked words are used to form query.}
\label{result3}
\resizebox{0.45\textwidth}{!}{%
\begin{tabular}{c|llll}
\hline

\hline
\multirow{3}{*}{\#Words} & \multicolumn{4}{c}{Ranking Method}\\

 & ${R}_{Orig}$ & $R_{Norm}$ & ${R}_{TFIDF}$  & ${R}_{IDF}$  \\
 \hline
 \multicolumn{5}{c}{New York Times Dataset}  \\
 \hline
	
5  &	0.0945 &	0.0463  &	\textbf{0.1363}	& 0.1187 \\
10 &	0.1161 &	0.0608 &	\textbf{0.1417} & 0.1291 \\
20 &	0.1256 &	0.0721	&   \textbf{0.1392} & 0.1321 \\
               
\hline

\multicolumn{5}{c}{Medline Dataset}  \\
\hline
5   &	0.1420  &	0.0202  &	 \textbf{0.1738}&   0.1578\\
10  &	0.1518  &	0.0289	&   \textbf{0.1612} &   0.1575\\
20  &   0.1498 	&   0.0372 &	 \textbf{0.1662} &   0.1642\\

\hline
\multicolumn{5}{c}{Amazon Product Reviews Dataset}  \\
 \hline
5  & 0.0231	& 0.0202 &	0.0208 & \textbf{0.0244}    \\
10 &   0.0195 & 	0.0154 & \textbf{0.0244} &	0.0236 \\
20 &	0.0258	& 0.0137 & 	\textbf{0.0279} &	0.0266  \\
        
\hline

\end{tabular}
}
\end{table}

\par Re-ranking words using  ${R}_{TFIDF}$ and ${R}_{IDF}$ consistently enhances retrieval performance compared to the default ranking (${R}_{Orig}$). ${R}_{TFIDF}$ produces the best results in the majority of configurations, the exception being when 5 words are used with the Amazon corpus where $R_{IDF}$ outperforms the other re-ranking methods. 
Re-ranking using $R_{Norm}$ is less effective than all the other rankings, including the default ranking. The relative performance of the four approaches is generally stable when the number of words used to form the query is varied and across the three datasets representing very different genres of text that were used in this experiment.

\par These results demonstrate that topic word re-ranking can produce words which are more effective for discriminating documents describing a particular topic from those which do not. The pattern of results is very similar to the crowdsourcing experiments suggesting that the re-rankings preferred by human subjects are those which are also useful within applications such as document retrieval.

\section{Evaluating Topic Representations}

\par This paper presented two novel methods for the evaluation of topic representations: a crowdsourcing experiment that relied on human judgments (Section \ref{experiment1}) and an automated evaluation based on an IR task (Section \ref{experiment2}). Although there are some differences between results using the two methods, the relative performance of the re-ranking methods explored in this paper are very similar. The correlations between results of the crowdsourcing experiment and IR evaluations are statistically significant for all three datasets (Pearson's $r$ varies between 0.81 and 0.90, $p<0.05$). This suggests that the automated evaluation approach presented in Section \ref{experiment2} is a useful tool for assessing the effectiveness of methods for word re-ranking with the advantage that results can be obtained more rapidly than methods that require human judgments. However, human judgments are recommended when performance is similar and automated evaluation should not be relied upon to make fine-grained distinctions between approaches, as is common for some tasks (e.g. Machine Translation \citep{Papineni2002}).

\section{Conclusion}
\label{conclusion}

\par We presented a study on word re-ranking methods designed to improve topic interpretability. Four methods were presented and assessed through two experiments. In the first experiment, participants on a crowdsourcing platform were asked to associate documents with related topics. In the second experiment, automated evaluation was based on a document retrieval task. 

\par Re-ranking the topic words was found to improve the interpretability of topics and therefore should be used as a post-processing step to improve topic representation. The most effective re-ranking schemes were those which combined information about the importance of words both within topics and their relative frequency in the entire corpus, thereby ensuring that less informative words are not used.

\bibliographystyle{chicago}
\bibliography{references}

\end{document}